\documentclass[a4paper]{article}
\usepackage[a4paper, top=2.5cm, bottom=2.5cm, left=2cm, right=2cm]{geometry}
\usepackage{tikz}
\usepackage{doi}
\usepackage{amsmath}
\usepackage{amssymb}
\usepackage{natbib}
\usepackage{subcaption}
\usepackage{hyperref}
\usepackage[nameinlink]{cleveref}

\providecommand{\keywords}[1]{\smallskip\noindent\textbf{Keywords:} #1}
\providecommand{\msc}[1]{%
  \smallskip\noindent{\footnotesize\textbf{Mathematics Subject Classification (2020):} #1}}

\title{\texttt{phepy}: Visual benchmarks and improvements for out-of-distribution detectors}

\author{%
  Felix Krumbiegel\footnotemark[1]
  \and Juniper Tyree\footnotemark[2]
  \and Michael Boy\footnotemark[2] \footnotemark[3]
  \and Petri Clusius\footnotemark[2]
  \and Andreas Rupp\footnotemark[1]
}

\begin{document}
\maketitle
\footnotetext[1]{\footnotesize Department of Mathematics, Saarland University, Saarland, Germany. \href{mailto:felix.krumbiegel@uni-saarland.de}{\texttt{felix.krumbiegel@uni-saarland.de}}}
\footnotetext[2]{\footnotesize Institute for Atmospheric and Earth System Research, University of Helsinki, Helsinki, Finland}
\footnotetext[3]{\footnotesize School of Engineering Sciences, LUT University, Lahti, Finland}

\abstract{
Applying machine learning to increasingly high-dimensional problems with sparse or biased training data increases the risk that a model is used on inputs outside its training domain. For such out-of-distribution (OOD) inputs, the model can no longer make valid predictions, and its error is potentially unbounded. Since testing OOD detection methods on real-world datasets is complicated, we design a benchmark for OOD detection, which includes three novel and easily-visualisable toy examples. These simple examples provide direct and intuitive insight into whether the detector is able to detect (1) linear and (2) non-linear concepts and (3) identify thin in-distribution (ID) subspaces (needles) within high-dimensional spaces (haystacks). We use our benchmark to evaluate the performance of various methods from the literature. Since tactile examples of OOD inputs may benefit OOD detection, we also review several simple methods to synthesise OOD inputs for supervised training. We introduce two improvements, $t$-poking and OOD sample weighting, to make supervised detectors more precise at the ID-OOD boundary. This is especially important when conflicts between real ID and synthetic OOD sample blur the decision boundary. Finally, we provide recommendations for constructing and applying OOD detectors in machine learning.}


\msc{62H30; 68T10.}

\section{Introduction} \label{sec:introduction}

In machine learning, the closed-world assumption is generally assumed to hold. That is, it is assumed that a model's \textbf{in-distribution} (ID) training data is drawn independently and identically distributed from the same distribution as the test data and the data on which the model is later applied \cite{ood-boundary-2021}. However, this assumption is broken when the model is applied to new \textbf{out-of-distribution} (OOD) data~\cite{generalized-ood-2024}, and the model's predictions can no longer be trusted. %
Detecting OOD inputs, thus, is a critical component of machine learning.

This paper explores two crucial aspects of OOD detection: \Cref{sec:ood-detection} presents an overview of existing methods for identifying novel inputs, and visualises  different methods on three toy examples for OOD detection. In the following sections we provide techniques to synthesise OOD inputs for supervised training methods. In \Cref{sec:ood-synthesis}, we introduce $t$-poking, and visualise the method using the same three toy examples, and in \Cref{sec:id-ood-weighting} we introduce weighting which we apply it to the SOSAA dataset, see~\cite{sosaa-2011,sosaa-data-2023}. %
We draw conclusions in \Cref{sec:conclusions}. 

Parts of this work follow along the lines of the master's thesis~\cite{surface-models-2023}.

\section{Toy-Comparisons of OOD Detection Methods and Related Work} \label{sec:ood-detection}

\noindent Evaluating OOD detectors is often difficult as the complexity of the data can obscure how a method performs. We introduce \href{https://github.com/juntyr/phepy}{\texttt{phepy}}, an open-source Python package to visually evaluate OOD detectors using three intuitive toy examples:
\begin{enumerate}
    \item Two groups of training points are scattered along a line in the 2D feature space. The target output variable only depends on the location along the line. In this example, points off the line are definitely OOD.
    \item The training points are scattered around the sine-displaced boundary of a circle, but none are inside it. The target output variable only depends on the location along the boundary. Again, points off the boundary are OOD.
    \item The training points are sampled from a 10-dimensional multivariate normal distribution but one of the features is set to a constant. This example tests whether an OOD detector can find a needle in a haystack.
\end{enumerate}%
In the following figures, each row showcases one detector. Its detection score is transformed into a CET-L20-colour-coded \cite{color-cet-2015} confidence level representing the percentage of validation data with greater or equal ID scores \cite{dime-detector-2021}. %
In each row, the first two plots depict a subset of the training points in unlabelled 2D feature space with black markers. For the third plot we instead show the \textit{distribution} of confidence values. On the $x$-axis we have the dimension of the value that is set to a constant (here marked as the white line). The y-axis now shows the confidence values. A point on the graph is then a tuple consisting of the value for the particular dimension and the confidence score. %

\begin{figure*}[!t]
    \centering
    \includegraphics[width=0.95\textwidth]{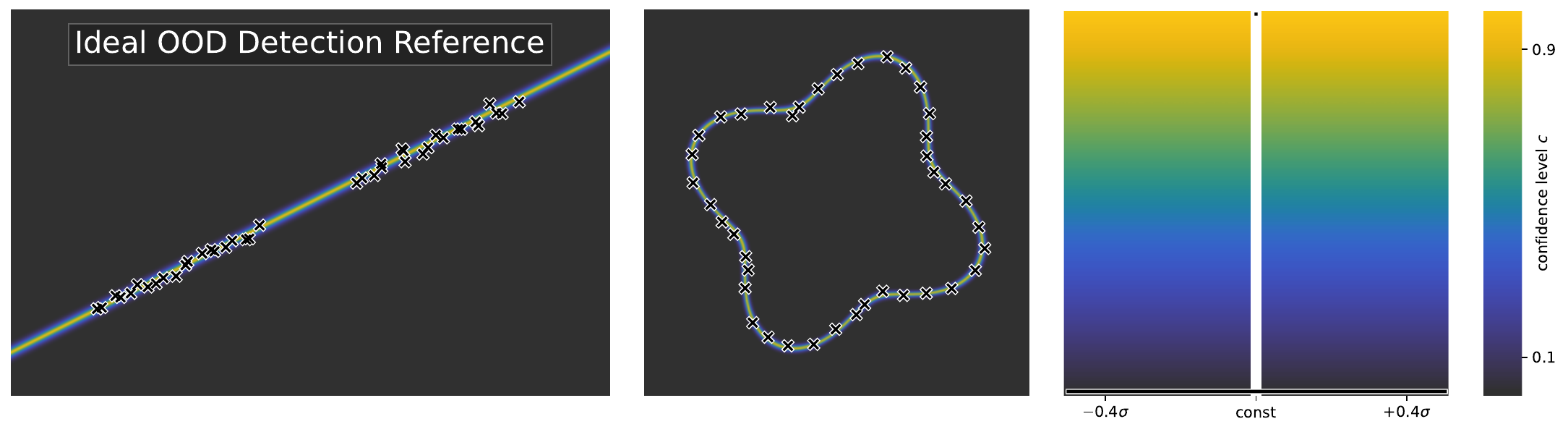}
    \caption{Visualisation of the reference OOD detector. Note that the reference is only one possible ideal detection method.}
    \label{fig:ideal-ood-detection}
\end{figure*}

\Cref{fig:ideal-ood-detection} portrays the reference OOD detection. In the first example, ID samples can be found on the whole line, which may be explained by assuming that our model may be able to interpolate the values from the two groups. For the transformed circle boundary (second panel) we observe that the whole circle line is ID and everything else is OOD. In the third panel the optimal detector scores only samples with the constant value (white column) highly confident, and everything else with low confidence. %

\begin{figure*}[!t]
    \centering
    \includegraphics[width=0.95\textwidth]{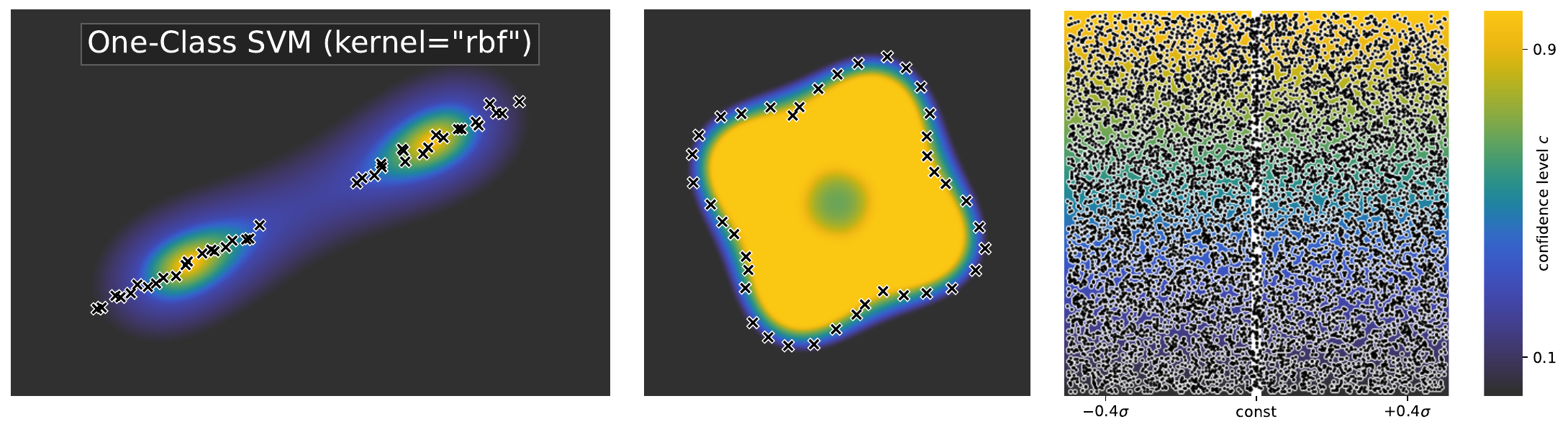}
    \caption{Visualisation of the one-class SVM.}
    \label{fig:svm-ood-detection}
\end{figure*}

The first OOD detector that is considered in \Cref{fig:svm-ood-detection} is the support vector machine (SVM) introduced by~\cite{support-vector-method-1999}. Classically the SVM classifies inputs into two data sets by separating them using a hyperplane. The method can be extended using non-linear kernel functions to separate the input sets in a transformed (potentially higher-dimensional) feature space. This work uses an RBF-kernel. %
This unsupervised method is available in \texttt{sklearn} \cite{scikit-learn-2011}. %
It generates two blurry confidence hills for the line example where some OOD inputs away from the line near the centre of the hills have higher confidence than some ID inputs that lie on the line but are `far' away from the centre of the hills. Moreover, it also completely fails for the circle, where it classifies the whole circle as ID, and for the haystack, where the confidence is uniformly distributed. %

\begin{figure*}[!t]
    \centering
    \includegraphics[width=0.95\textwidth]{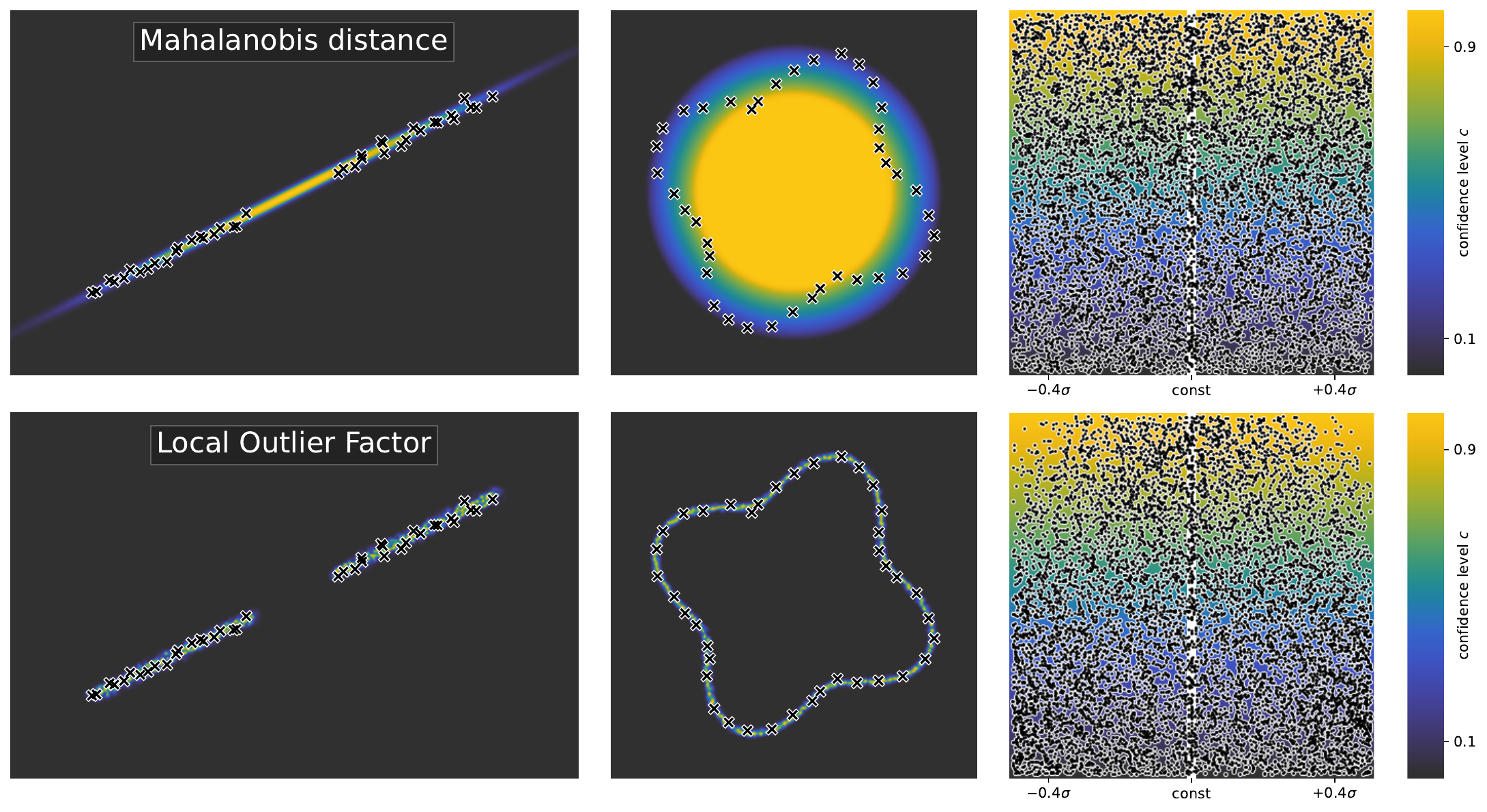}
    \caption{Visualisation of the Mahalanobis distance and the Local Outlier Factor as OOD detectors.}
    \label{fig:md-lof-if-ood-detection}
\end{figure*}

Next, \Cref{fig:md-lof-if-ood-detection} compares distance- and isolation-based detectors. %
The first row shows detection results of the Mahalanobis distance by~\cite{mahalanobis-outlier-2000}. The idea of the Mahalanobis distance is to find a multivariate normal distribution based on the training ID inputs and measure the deviation of novel inputs. Since the feature distances on high-dimensional inputs are vulnerable to noise and difficult to interpret, the distances are often computed on lower-dimensional embeddings, e.g. created by the upstream levels of a neural network \cite{dime-detector-2021}. %
Unsurprisingly, the Mahalanobis distance works very well on the line example and fails on the circle line as it cannot be well represented by a normal distribution. Further, the Mahalanobis distance does not perform well in higher dimensions which explains the failure on the haystack example. %
The second row shows OOD detection using the Local Outlier Factor (LOF) \cite{lof-outlier-2000}, which estimates the density of ID training data around the input and assigns a high confidence score to inputs in regions with a high density. %
For the two low-dimensional examples, the line and the circle line, the LOF performs very well, even if it detects the line too conservatively in the sense that it classifies two separate sets instead of interpolating between. {For the haystack, the LOF shows an improved detection compared to the previous examples, however, it still fails to properly classify OOD inputs}. %

\begin{figure*}[!t]
    \centering
    \includegraphics[width=0.95\textwidth]{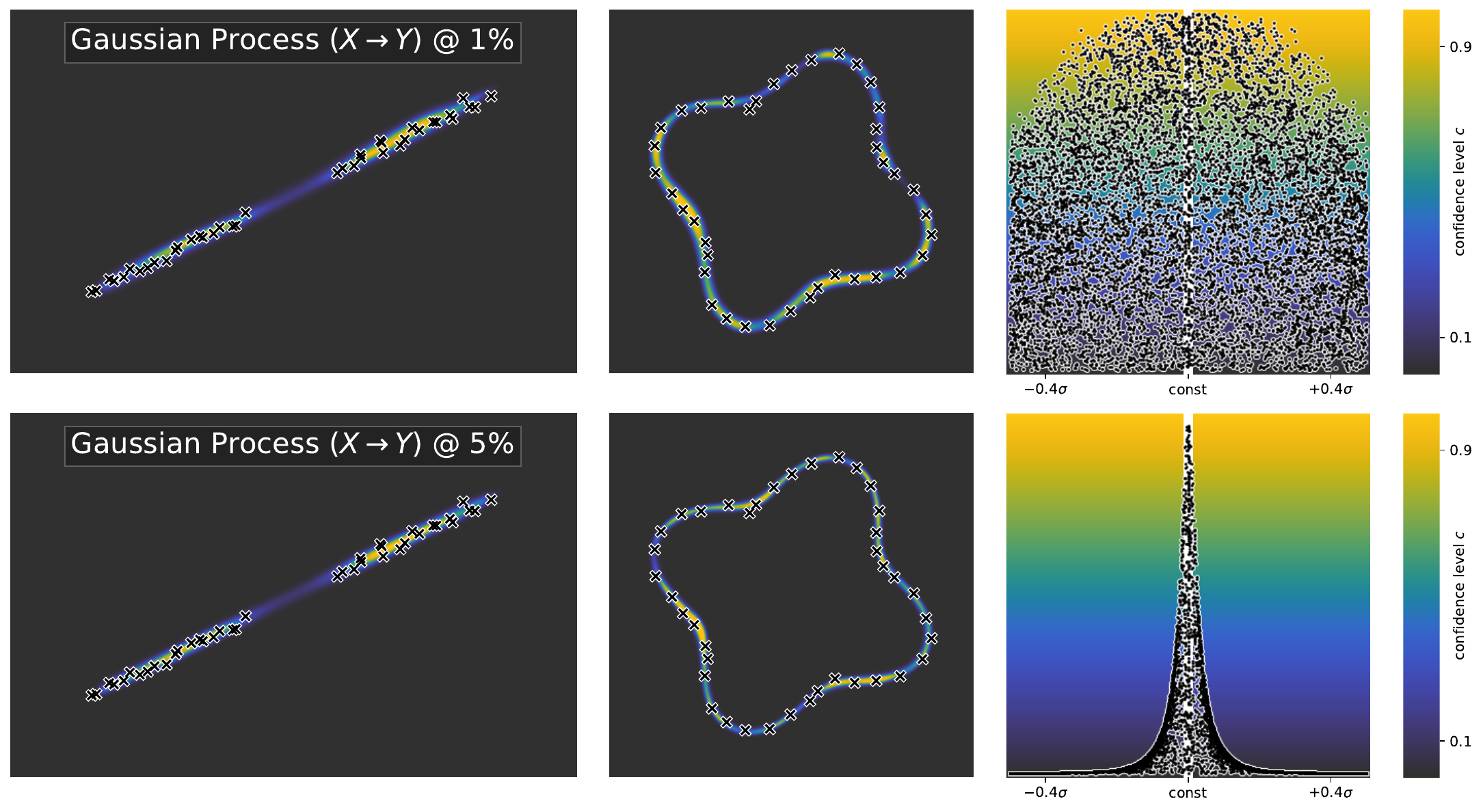}
    \caption{Visualisation of the Random Forest and two Gaussian Processes as OOD detectors.}
    \label{fig:rf-gp-ood-detection}
\end{figure*}

Many OOD inputs are semantically valid inputs missing from the training data, which are, thus, only OOD due to incomplete knowledge about the parameter space. %
Here, we present an example for an uncertainty-based OOD detectors in \Cref{fig:rf-gp-ood-detection}. %
The two rows show the detection using a Gaussian process (GP) \cite{gaussian-process-2005}, which fits a multivariate normal distribution to the training data. %
We observe that this detector clearly out-performs all previous approaches in the three test cases. However, GPs come with an immense cost of~$\mathcal{O}(N^3)$, and subsampling is required for the high-dimensional haystack example, where a subsampling of at least~$5\%$ is necessary. 

\begin{figure*}[!t]
    \centering
    \includegraphics[width=0.95\textwidth]{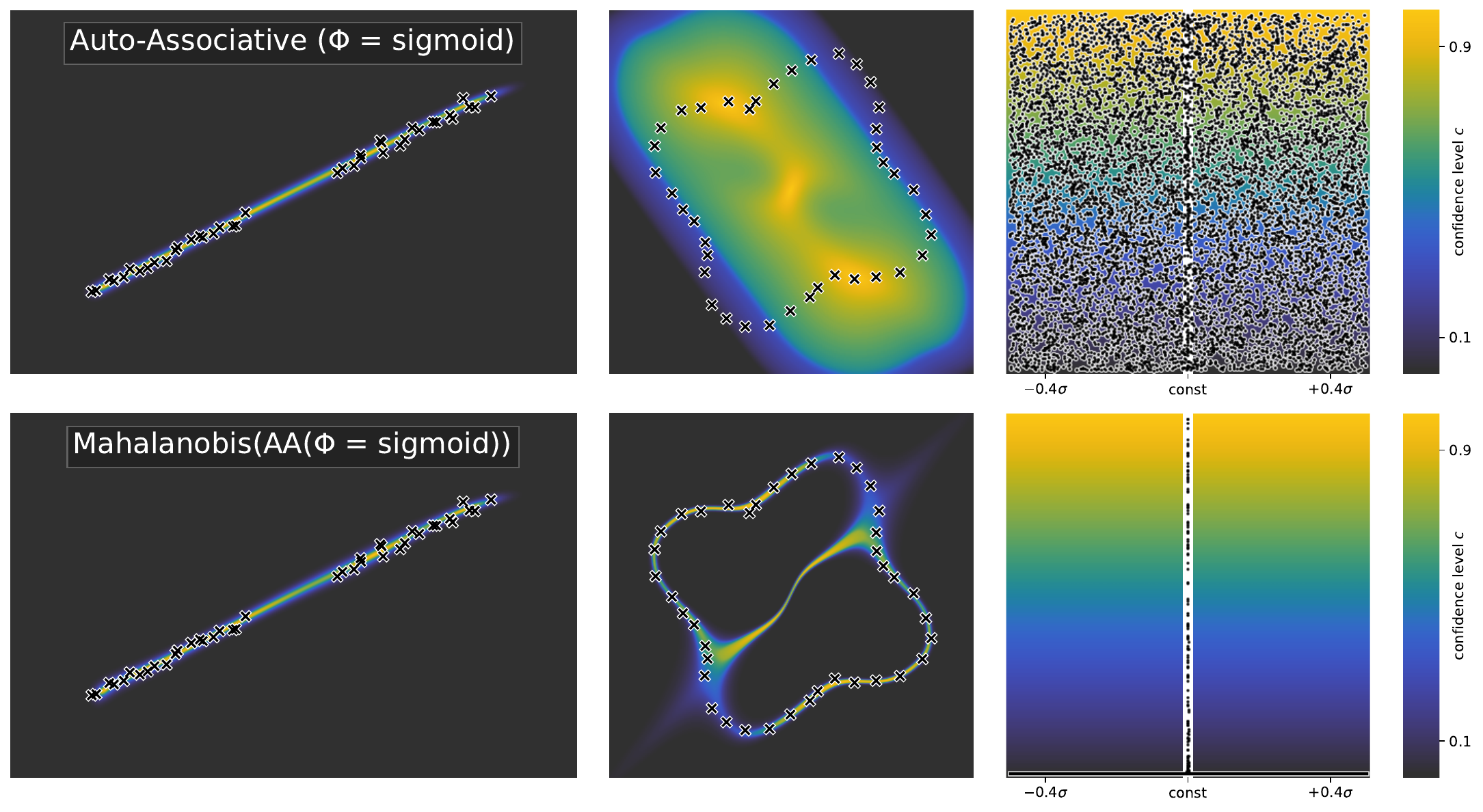}
    \caption{Visualisation of auto-encoders (sharpened by the Mahalanobis distance) as auto-associative OOD detectors.}
    \label{fig:aa-ood-detection}
\end{figure*}

\begin{figure*}[!t]
    \centering
    \includegraphics[width=0.95\textwidth]{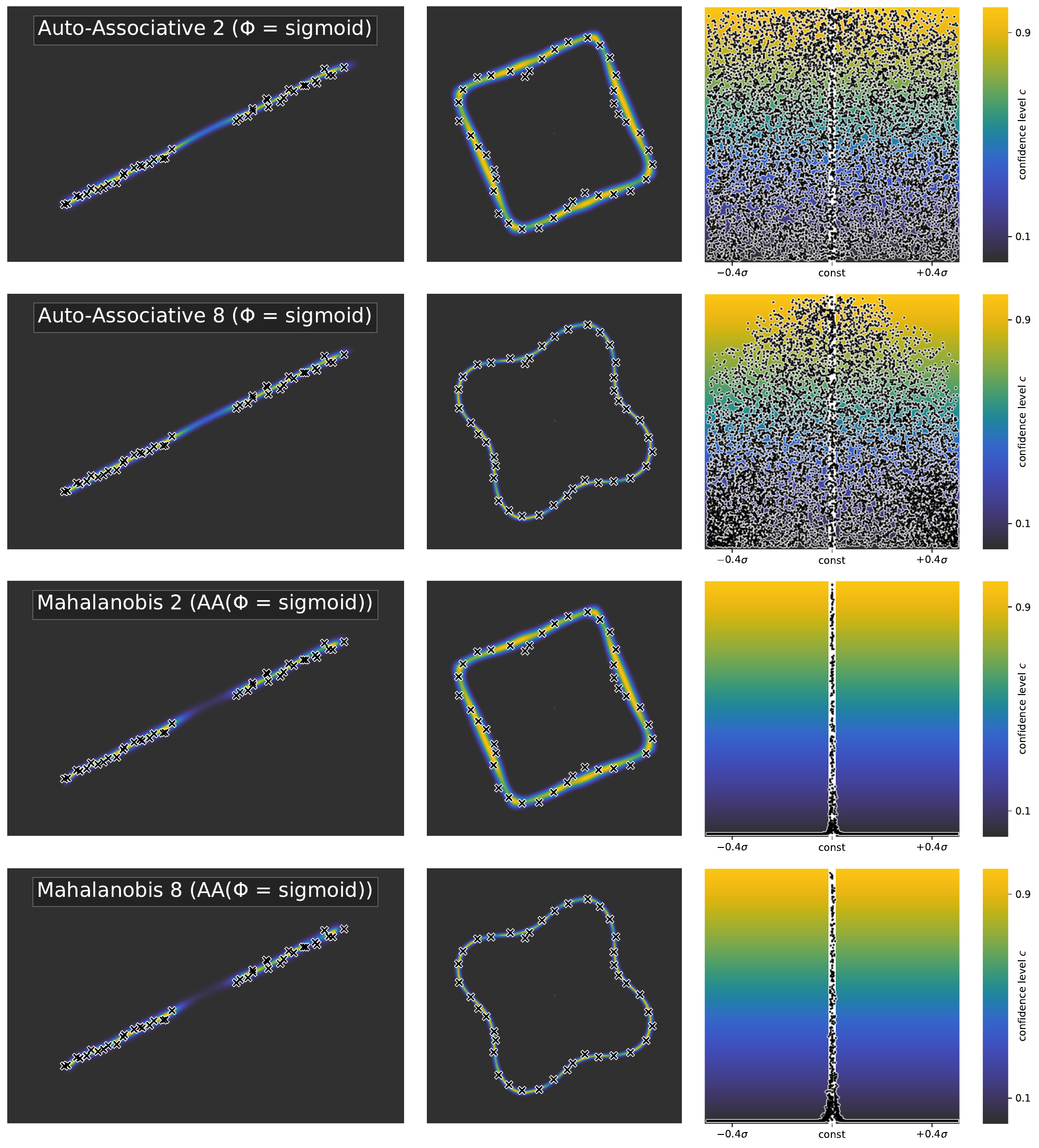}
    \caption{Visualisation of auto-encoders (sharpened by the Mahalanobis distance) as auto-associative OOD detectors with channel sizes 2 and 8, respectively.}
    \label{fig:aa-ood-detection-2}
\end{figure*}

Finally, in \Cref{fig:aa-ood-detection} we present a selection of auto-associative methods. %
The figure shows detection using simple auto-encoder multilayer perceptrons, which are constructed with $16 \times 1 \times 16$ hidden layers using sigmoid activation functions. The difference in the auto-encoders lies in the loss function that is minimized, i.e., for the method in the first row, the standard mean squared error is used. Except for the line example the performance is suboptimal. %
This suboptimal behaviour of the mean squared error motivates to use the Mahalanobis distance of the errors instead, which greatly improves the OOD detection for all three examples, as shown in the second row. Specifically, for the haystack example this OOD detector behaves similarly to the optimal OOD detector from \Cref{fig:ideal-ood-detection}. %
However, for the circle the detection is not sufficient. %
Note, that this follows from the architectural choice for the auto-encoder. \Cref{fig:aa-ood-detection-2} shows OOD detection for auto-encoders where the channel sizes are increased, i.e., we use $16 \times 2 \times 16$, and $16 \times 8 \times 16$ hidden layers, respectively. For the line and circle example increasing the channel size yields much better results for both auto-encoder, and Mahalanobis distance auto-encoder. Solely for the haystack example using Mahalanobis, the channel with size 1 is best even if larger channels are only marginally worse. This may be explained by the construction of this example as a truly one-dimensional ID data set in a higher-dimensional space. %

\section{Toy-Comparisons of OOD Synthesis Methods} \label{sec:ood-synthesis}

\noindent \Cref{sec:ood-detection} has examined several unsupervised OOD detection methods, which are trained using only ID samples. However, tasks such as providing meaningful OOD confidence scores benefit from being trained with both ID and OOD samples. This section evaluates several simple approaches to synthesising OOD inputs given the ID inputs. First, we train a multi-layer perceptron classifier on the ID and synthetic OOD inputs using the default settings in \texttt{sklearn} \cite{scikit-learn-2011} and take its prediction probability as its confidence. We again use the three toy examples from \Cref{sec:ood-detection} to evaluate the final detector's performance, which reflects the synthesis method's quality. To visualise the OOD synthesiser's performance, the line and circle plots now show synthetic OOD points as white dots. 

Sampling random noise is a commonly used technique to synthesise OOD inputs \cite{noise-contrastive-uq-2019, ood-training-2017, ood-boundary-2021}. %
However, OOD and ID inputs are not prevented from overlapping, and the detector thus fails to separate OOD inputs in the haystack. %

\begin{figure*}[!t]
    \centering
    \includegraphics[width=0.95\textwidth]{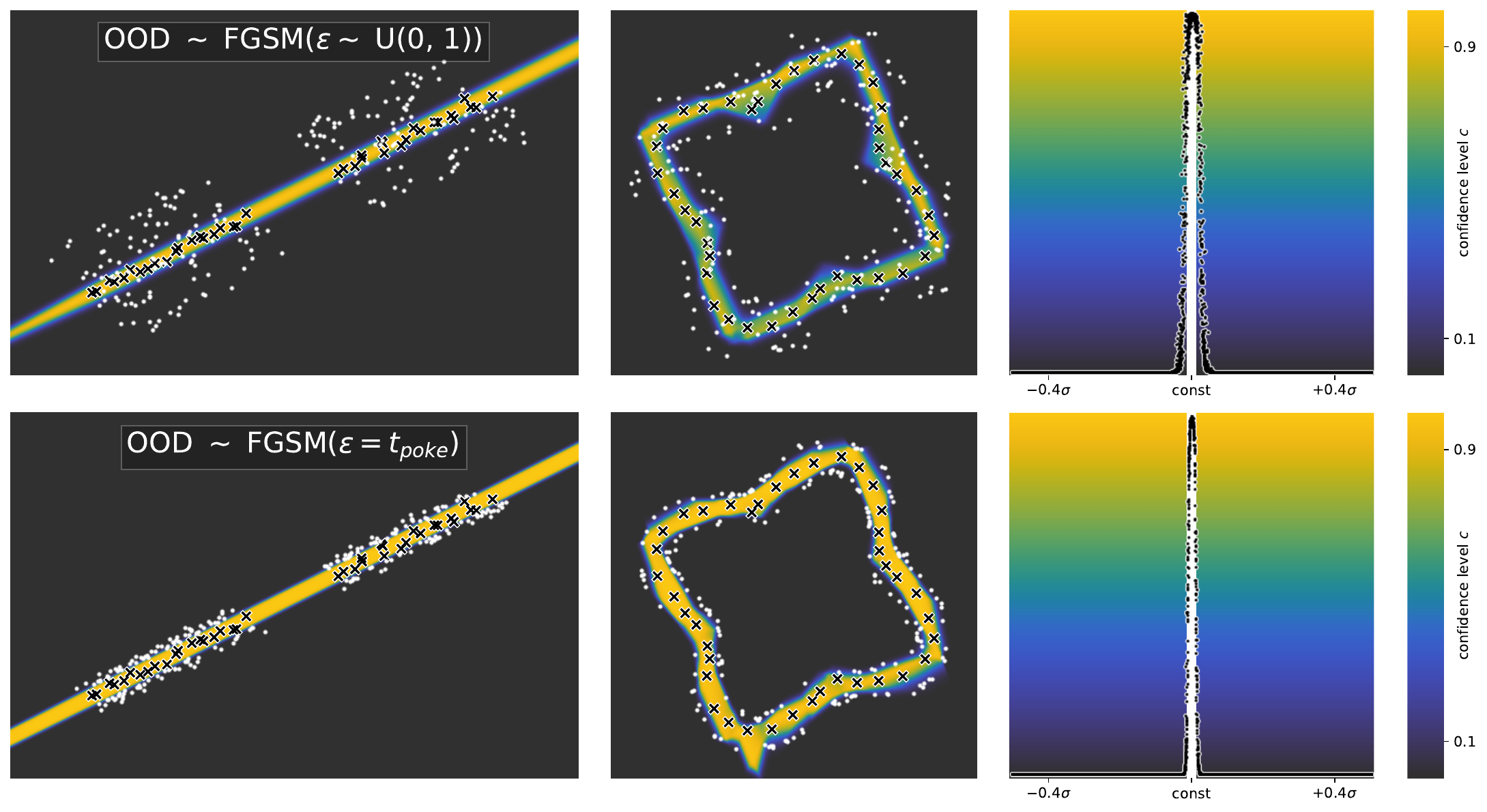}
    \caption{Comparison of supervised OOD detectors using inputs synthesised with FGSM with uniformly sampled and $t$-poking-determined step lengths.}
    \label{fig:fgsm-eps-u01-tpoke-ood-synthesis}
\end{figure*}

Adversarial methods provide a more targeted approach to generating OOD inputs. We test the Fast Gradient Sign Method (FGSM) \cite{fast-gradient-2014}, which perturbs inputs to maximise an error metric. We first use kernel density estimation \cite{kde-1956} with a Gaussian kernel to synthesise new ID inputs. Next, the FGSM perturbs them to increase the reconstruction error of the auto-associative reference detector from \Cref{sec:ood-detection}. As the FGSM thus has access to the ideal error gradients, we evaluate its best-case performance. %
Using a constant perturbation step size highly depends on the problem and may in too conservative or too crisp OOD boundaries. %
Thus, in \Cref{fig:fgsm-eps-u01-tpoke-ood-synthesis}, we show the result when sampling the step size from a distribution, and observe that this procedure produces tight but also blurry boundaries as some ID-OOD overlap may occur. %
This indicates that it might be beneficial to choose an FGSM step length which tightens the OOD boundary without impacting the confidence in ID inputs. Since manual tuning would be laborious, we introduce $t$-poking, which sets the step size to a new parameter or distribution $t$. After every batch, epoch or training cycle, the detector is evaluated using criteria such as ``the mean confidence score for ID inputs should not fall below $0.95$''. If the evaluation fails, $t$ is multiplied by a back-off factor $>1$. Otherwise, $t$ is multiplied by a poking factor $<1$. This factor is gradually adjusted to ensure that $t$ converges. The second row in \Cref{fig:fgsm-eps-u01-tpoke-ood-synthesis} shows that $t$-poking finds a balanced step size and pushes the OOD samples very close to the ID boundary. %

\section{Weighting Synthetic OOD Samples in Detector Training} \label{sec:id-ood-weighting}

\begin{figure*}[!t]
    \centering
    \includegraphics[width=0.95\textwidth]{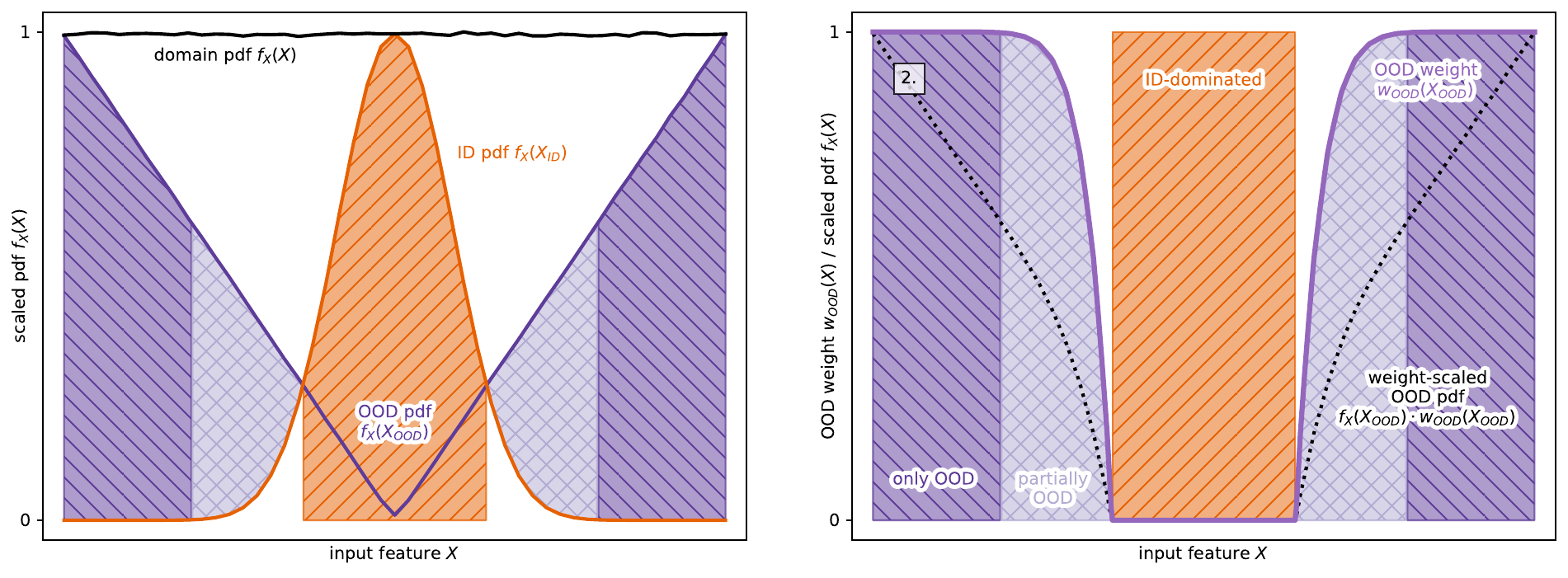}
    \caption{%
     Where ID samples dominate the input domain, shown in orange stripes, OOD samples are given zero weight to avoid ID-OOD overlap. In contrast, areas without ID samples, shown in purple, are assigned full OOD weight.}
    \label{fig:id-ood-weights}
\end{figure*}

\noindent Overlapping ID and synthetic OOD points cannot be solved with clever OOD synthesis alone. For instance, applying an OOD synthesiser to an entirely ID parameter space would still produce synthetic OOD samples, even though they are guaranteed to be ID. This issue motivates us to define a weighting function $w_{\text{OOD}}(X) \in [0.0; 1.0]$ that weighs down OOD inputs that clash with ID training data. We assume that there is a distribution $\mathbb{P}_{X}$ of semantically valid input features that is a superset of the ID and OOD distributions. Since the synthetic OOD inputs may contain some misclassified ID inputs, we want to prioritise ID samples over synthetic OOD samples. Thus, after the ID inputs have been drawn from $\mathbb{P}_{X}$, they partially block off parts of the distribution. When synthetic OOD inputs are generated, samples that fall within ID areas are blocked and weighted down. \Cref{fig:id-ood-weights} shows a simple example which we use to explain how we design the OOD sample weight factor $w_{\text{OOD}}(X)$.

We first need to calculate the probability density functions (PDF) of the input domain distribution $\mathbb{P}_X$, the in-distribution $\mathbb{P}_{X_{\text{ID}}}$ and the out-of-distribution $\mathbb{P}_{X_{\text{OOD}}}$. If their analytical forms are unknown, they can be proxied, e.g. with distribution fitting or a lower-dimensional auto-associative reconstruction error. %
Note that some of the areas have been sampled twice, i.e. the ID PDF and the OOD PDF overlap. Since we prioritise ID samples, we ignore the area below the OOD PDF wherever it overlaps with the area below the ID PDF curve.

Finally, we can calculate the weighting factor for ID and OOD samples. For any input $x \in X$, we look at the fraction of the area under the combined sampling curves, the maximum of $f_{\text{ID}}(X)$ and $f_{\text{OOD}}(X)$, that is taken up by ID to calculate the ID sample weight. The OOD sample weight $w_{\text{OOD}}(X)$ then is its complement:
\begin{equation*}
    w_{\text{OOD}}(X) = 1 - \frac{f_{\text{ID}}(X)}{\max(f_{\text{ID}}(X), f_{\text{OOD}}(X))}
\end{equation*}%
We now evaluate the impact of applying the OOD sample weighting scheme. For simplicity, we assume that $\mathbb{P}_X = \mathbb{P}_{X_{\text{OOD}}}$ for the uniform and FGSM methods, i.e. that OOD samples are drawn with some uniformity from the input domain distribution. %
Further, we show the impact on a real-world data set, the SOSAA data set consisting of aerosol trajectories. The data is based on the SOSAA model~\cite{sosaa-2011,sosaa-data-2023}, and contains measurements made at the SMEAR II station at Hyytiälä~\cite{smear-2013}. %
We have given the different evaluation datasets: 
\begin{enumerate}
    \item ID training and validation data in dark blue. This should have full confidence.
    \item ID test data in light blue. This should result in high confidence.
    \item In purple, we show approximation of the ID data by a multivariate normal distribution that uses the empirical mean and covariance of the training data set. This approximation may be classified ID or OOD.
    \item Temporally adjacent data to the training data is shown in green. Should score high confidence given the relative stability of the trajectories.
    \item In yellow, we have five other trajectories classified either as ID or OOD.
    \item This dataset interpolates between the approximation of the ID dataset and a definitely-OOD normal distribution and is shown in orange.
    \item The last dataset portrayed in red show samples from a multivariate normal distribution~$\mathcal{N}(0, I)$ and should be assigned zero confidence since the features of the SOSAA dataset are not independent. 
\end{enumerate}

\begin{figure*}[!t]
    \centering
    \begin{subfigure}[!t]{.5\textwidth}
        \centering
        \includegraphics[width=0.95\textwidth]{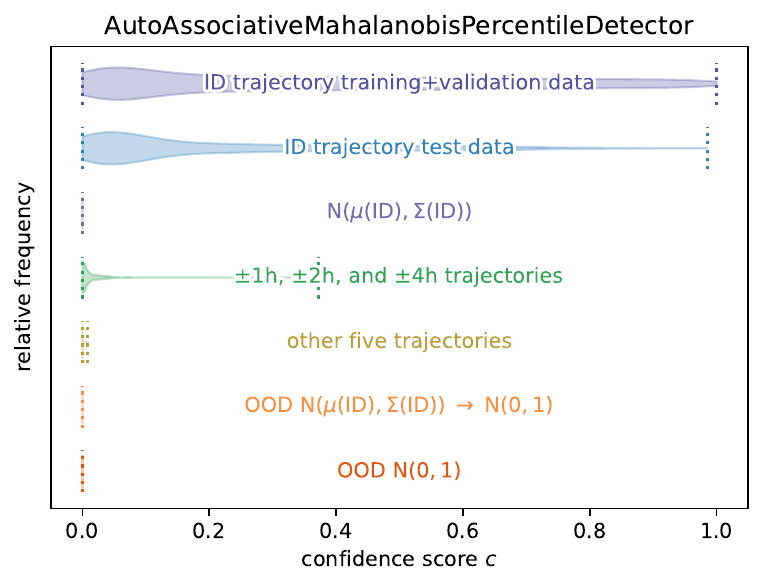}
    \end{subfigure}%
    \hfill%
    \begin{subfigure}[!t]{.5\textwidth}
        \centering
        \includegraphics[width=0.95\textwidth]{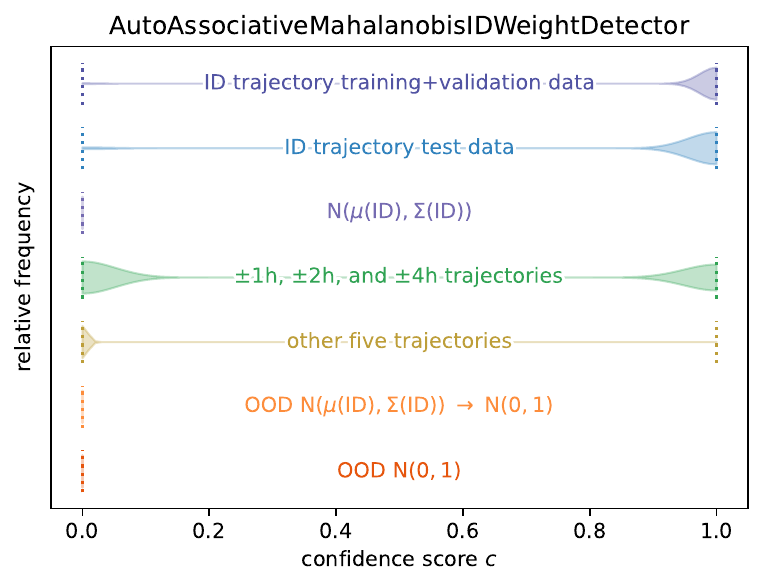}
    \end{subfigure}

    \begin{subfigure}[!t]{.5\textwidth}
        \centering
        \includegraphics[width=0.95\textwidth]{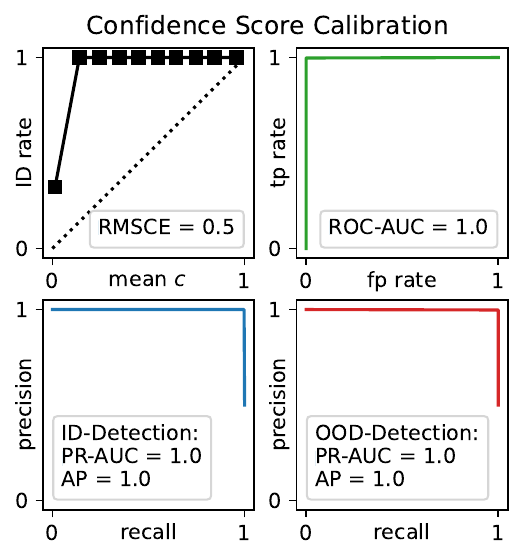}
    \end{subfigure}%
    \hfill%
    \begin{subfigure}[!t]{.5\textwidth}
        \centering
        \includegraphics[width=0.95\textwidth]{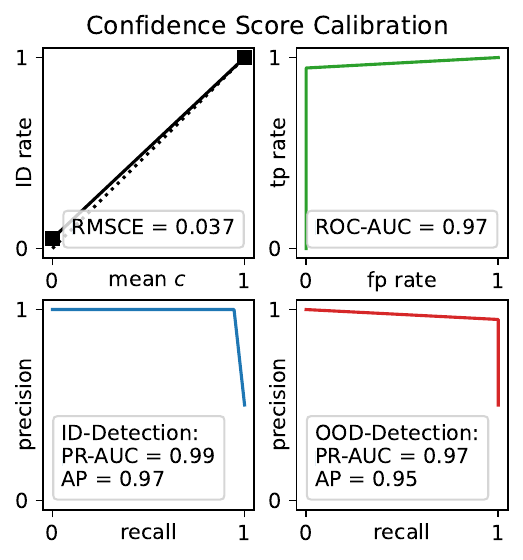}
    \end{subfigure}%
    \caption{Confidence score distributions of the auto-encoder with the Mahalanobis distance for the reconstruction error from the previous section. The right column shows the detector employing weighted synthesized OOD inputs.}
    \label{fig:sosaa-ood-synthesis}
\end{figure*}

In \Cref{fig:sosaa-ood-synthesis} on the left we have the confidence values, where the thickness indicates the distribution of confidence values and the $x$-axis shows the confidence score. %
We observe that the confidence scores align quite well with the expected confidence scores for the examples, specifically, the detector scores clearly OOD inputs with very low confidence. However, the auto-encoder seems to be too underconfident even for clearly ID inputs. Note that being underconfident is generally preferred since being overconfident in wrong data leads users to trust faulty results. %
This drawback can be somewhat remedied when training the auto-encoder with synthesised OOD inputs using weighting. The plot on the right shows that the detector still has near zero confidence for OOD inputs, and scores most data points as either high confidence or low confidence where only a fraction of inputs has confidence scores in between. %

The four tables below evaluate quantitatively how well the scoring function distinguishes between ID and OOD inputs. For the scoring we consider only the test and definitely-OOD data sets as there we have true labels given. %
The first table shows how well the confidence scores are calibrated. There, we split the dataset into ten bins based on their confidence score. Ideally the bins should lie on the diagonal as the ID rate should grow proportionally with the confidence score. If this is not the case, staying above the diagonal is preferred as being underconfident is better than being overconfident. %
Here, we observe that the auto-encoder with the Mahalanobis distance is quite underconfident overall, which motivates the use of OOD synthesis with weighting. On the right-hand side the first plot shows now almost perfect calibration, which is reflected with the RMSCE score. %
The top-right plot shows the ROC-AUC which compares the false-positive with the true-positive rate for the ID inputs. The ideal plot should hug the top-left corner and the AUC should be (close to) 1. %
The two bottom plots show the precision for the ID and OOD detection rate, that is the recall (true positive rate) is plotted against the precision, which is the percentage of true ID/OOD detection. This plot should cover the top-right corner and the AUC should ideally be 1. 

We observe in \Cref{fig:sosaa-ood-synthesis} that the auto-encoder without weighting is already very good at scoring the correct OOD and ID inputs, as the AUC for the ROC and recall are~1. However, the calibration leaves room for improvement, and employing the weighting greatly improves this calibration with only a minor decrease in the ROC and recall AUC. 

\section{Conclusions} \label{sec:conclusions}

This paper has explored OOD detection, a critical component of safe machine learning, from several perspectives. First, \Cref{sec:ood-detection} introduced \textbf{three toy examples} to evaluate OOD detection methods. These toy examples are designed to visualise the performance of a particular method using the open-source \href{https://github.com/juntyr/phepy}{\texttt{phepy}} Python package, with which all plots for this paper were produced, to help others use these toy examples as a standard comparison test in the future. %
In \Cref{sec:ood-detection}, we also compared a variety of unsupervised detectors with the three toy examples. We thus learned that \textbf{auto-associative methods} with the Mahalanobis distance and \textbf{Gaussian Processes} produce the most generalisable results, though the latter come with high computational costs. %
\Cref{sec:ood-synthesis} has examined synthesising OOD inputs to train a supervised OOD classifier. Our experiments demonstrate the power and shortfalls of simple methods such as randomly sampling OOD inputs or using the \textbf{fast gradient sign method} to generate them adversarially. Since the supervised detectors produce crisper but more conservative OOD detection boundaries, we also briefly introduced two improvements, $t$-poking and OOD sample weighting, to tighten this boundary.

\section*{Acknowledgements}

The authors wish to acknowledge the IT for Science Group at the University of Helsinki for computational resources. %
Parts of this work were conducted during F.~Krumbiegel's and A.~Rupp's stay at the Hausdorff Research Institute for Mathematics funded by the Deutsche Forschungsgemeinschaft (DFG, German Research Foundation) under Germany's Excellence Strategy -- EXC-2047/2 -- 390685813. 
This work has been supported by the Deutsche Forschungsgemeinschaft (DFG, German Research Foundation) -- 577175348. %
We thank the reviewer for their helpful comment on the setup of the auto-encoder for the OOD detection which helped improve the presentation. 



\begin{thebibliography} {99}


\bibitem[\protect\astroncite{Boy et~al.}{2023}]{sosaa-data-2023}
Boy, M., Chen, D., Clusius, P., Gierens, R., Smolander, S., Taipale, D., Tyree, J., Zhou, L., and Zhou, P. (2023).
\newblock SOSAA (msc-tcm version) \doi{10.5281/zenodo.7867027}.

\bibitem[\protect\astroncite{Boy et~al.}{2011}]{sosaa-2011}
Boy, M., Sogachev, A., Lauros, J., Zhou, L., Guenther, A., and Smolander, S. (2011).
\newblock SOSA – a new model to simulate the concentrations of organic vapours and sulphuric acid inside the ABL – Part 1: Model description and initial evaluation.
\newblock {\em Atmospheric Chemistry and Physics}, 11(1):43--51.

\bibitem[\protect\astroncite{Breunig et~al.}{2000}]{lof-outlier-2000}
Breunig, M.~M., Kriegel, H.-P., Ng, R.~T., and Sander, J. (2000).
\newblock Lof: Identifying density-based local outliers.
\newblock In {\em Proceedings of SIGMOD'00}, pages 93--104.

\bibitem[\protect\astroncite{Goodfellow et~al.}{2014}]{fast-gradient-2014}
Goodfellow, I.~J., Shlens, J., and Szegedy, C. (2014).
\newblock Explaining and harnessing adversarial examples.

\bibitem[\protect\astroncite{Hafner et~al.}{2019}]{noise-contrastive-uq-2019}
Hafner, D., Tran, D., Lillicrap, T., Irpan, A., and Davidson, J. (2019).
\newblock Noise contrastive priors for functional uncertainty.

\bibitem[\protect\astroncite{Hari et~al.}{2013}]{smear-2013}
Hari, P., Nikinmaa, E., Pohja, T., Siivola, E., B{\"a}ck, J., Vesala, T., and Kulmala, M. (2013).
\newblock {\em Station for Measuring Ecosystem-Atmosphere Relations: SMEAR}.
\newblock Springer Netherlands, Dordrecht.

\bibitem[\protect\astroncite{Kim}{2000}]{mahalanobis-outlier-2000}
Kim, M.~G. (2000).
\newblock Multivariate outliers and decompositions of mahalanobis distance.
\newblock {\em Communications in Statistics -- Theory and Methods}, 29(7):1511--1526.

\bibitem[\protect\astroncite{Kovesi}{2015}]{color-cet-2015}
Kovesi, P. (2015).
\newblock Good colour maps: How to design them.

\bibitem[\protect\astroncite{Lee et~al.}{2017}]{ood-training-2017}
Lee, K., Lee, H., Lee, K., and Shin, J. (2017).
\newblock Training confidence-calibrated classifiers for detecting out-of-distribution samples.

\bibitem[\protect\astroncite{Pedregosa et~al.}{2011}]{scikit-learn-2011}
Pedregosa, F., Varoquaux, G., Gramfort, A., Michel, V., Thirion, B., Grisel, O., Blondel, M., Prettenhofer, P., Weiss, R., Dubourg, V., Vanderplas, J., Passos, A., Cournapeau, D., Brucher, M., Perrot, M., and Duchesnay, E. (2011).
\newblock Scikit-learn: Machine learning in python.
\newblock {\em JMLR}, 12(85):2825--2830.

\bibitem[\protect\astroncite{Pei et~al.}{2021}]{ood-boundary-2021}
Pei, S., Zhang, X., Fan, B., and Meng, G. (2021).
\newblock Out-of-distribution detection with boundary aware learning.

\bibitem[\protect\astroncite{Rasmussen and Williams}{2005}]{gaussian-process-2005}
Rasmussen, C.~E. and Williams, C. K.~I. (2005).
\newblock {\em Gaussian Processes for Machine Learning}.
\newblock The MIT Press.

\bibitem[\protect\astroncite{Rosenblatt}{1956}]{kde-1956}
Rosenblatt, M. (1956).
\newblock {Remarks on Some Nonparametric Estimates of a Density Function}.
\newblock {\em The Annals of Mathematical Statistics}, 27(3):832--837.

\bibitem[\protect\astroncite{Sch\"{o}lkopf et~al.}{1999}]{support-vector-method-1999}
Sch\"{o}lkopf, B., Williamson, R.~C., Smola, A., Shawe-Taylor, J., and Platt, J. (1999).
\newblock Support vector method for novelty detection.
\newblock In Solla, S., Leen, T., and M\"{u}ller, K., editors, {\em Advances in Neural Information Processing Systems}, volume~12. MIT Press.

\bibitem[\protect\astroncite{Sjögren and Trygg}{2021}]{dime-detector-2021}
Sjögren, R. and Trygg, J. (2021).
\newblock Out-of-distribution example detection in deep neural networks using distance to modelled embedding.

\bibitem[\protect\astroncite{Tyree}{2023}]{surface-models-2023}
Tyree, J. (2023).
\newblock Prudent Response Surface Models : Exploring a Framework for Approximating Simulations with Confidence and Certainty

\bibitem[\protect\astroncite{Yang et~al.}{2024}]{generalized-ood-2024}
Yang, J., Zhou, K., Li, Y., and Liu, Z. (2024).
\newblock Generalized {Out}-of-{Distribution} {Detection}: {A} {Survey}.
\newblock {\em International Journal of Computer Vision}, 132(12):5635--5662.

	
\end{thebibliography}
\end{document}